\documentclass[runningheads]{llncs}

\usepackage{eccv}

\usepackage{eccvabbrv}

\usepackage{graphicx}
\usepackage{booktabs}

\usepackage[accsupp]{axessibility}  % Improves PDF readability for those with disabilities.

\usepackage{hyperref}

\usepackage{orcidlink}

\usepackage{listings}
\usepackage{array}

\begin{document}

\title{Periocular Soft Biometrics: A Survey and Applications to Multimedia Forensics and Disinformation Detection}

\titlerunning{Demographic Estimation from Periocular Images}

\author{Fernando Alonso-Fernandez\orcidlink{0000-0002-1400-346X} \and
Kevin Hernandez-Diaz\orcidlink{0000-0002-9696-7843} \and
Josef Bigun\orcidlink{0000-0002-4929-1262}}

\authorrunning{F.~Alonso-Fernandez et al.}

\institute{School of Information Technology (ITE), Halmstad University, Sweden \\
\email{\{feralo, kevin.hernandez-diaz, josef.bigun\}@hh.se}}

\maketitle

\begin{abstract}
Soft-biometric attributes such as gender, age, and ethnicity provide
valuable ancillary evidence when full identity recognition is not feasible, 
supporting applications in forensic investigation, identity verification, 
surveillance, or detection of synthetic and manipulated media.
Among biometric modalities, the periocular region is a robust source of soft-biometric cues, as it often remains visible when other parts of the face are occluded, a frequent condition in forensic evidence and surveillance footage, and can be captured across a wide range of acquisition conditions.
In this paper, we provide a survey of 
demographic attribute estimation from periocular images, covering publicly
available datasets, methodological trends from handcrafted descriptors to
deep learning architectures, and the state of the art in gender, age, and ethnicity prediction.
We discuss use cases relevant to multimedia forensics and disinformation-detection applications, including demographic filtering in surveillance footage, age verification, and the detection of demographic inconsistencies in synthetic data.
We also highlight open challenges, including
dataset bias, cross-domain generalisation, fairness, ethical aspects, and the lack of forensic-oriented benchmarks.

\end{abstract}

%% ===========================================================
\section{Introduction}
\label{sec:intro}
%% ===========================================================
 
Face recognition is now embedded in mobile devices, border-control
systems, and large-scale surveillance infrastructures. However, recognition
performance degrades substantially under unconstrained conditions due to
occlusion, pose variation, illumination change, and low image quality \cite{2026PAMI_50yrfacerecognition_jain}.  
This limitation is especially relevant in forensic investigations, which frequently rely on partial, occluded, or low-quality data from surveillance cameras, crime-scene imagery, or deliberately concealed faces, for example, by offenders attempting to avoid identification. In such cases, full-face analysis may be infeasible.
Partial faces are also common in controlled or semi-controlled contexts involving face coverings, such as masks, protective or cultural gear, and digital filters on social media \cite{Alonso24computers_periSOA}.

\begin{figure}[t]
  \centering
  % ---------------------------------------------------------------
  % FIGURE SUGGESTION: Use a modified version of Fig.1 from the
  % journal paper (the labelled eye image). To avoid identical
  % figures in the two publications, we recommend rendering a
  % slightly different crop or annotation style -- e.g. using only
  % a schematic diagram with named anatomical parts rather than the
  % photograph -- or replacing the photograph with a freely-licensed
  % eye image from a public dataset (e.g. UBIRIS or ND-Iris-0405).
  % Placeholder command below; replace with actual figure file.
  % ---------------------------------------------------------------
  %\includegraphics[width=0.4\textwidth]{C1_S1_I7.png}
  \includegraphics[width=0.98\textwidth]{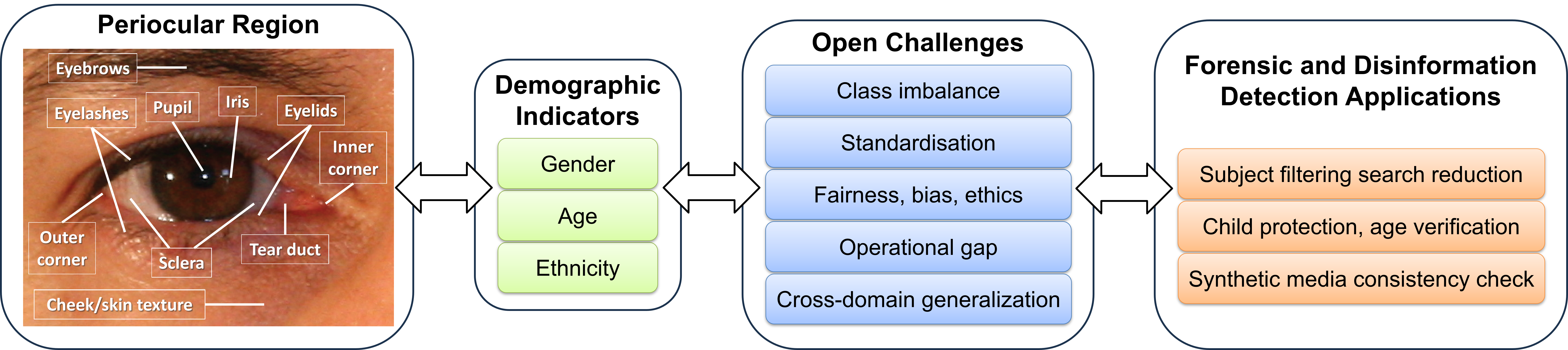}
  \caption{From left to right: $i$) anatomical parts of the periocular region (the periocular modality uses the full visible area around the eye, not just the iris region, image from \cite{[Alonso16]}); $ii$) demographic indicators; $iii$) limitations and open challenges in periocular soft biometrics research; and $iv$) applications of periocular soft biometrics to forensic and disinformation detection applications.}
  \label{fig:eye_anatomy}
\end{figure}

In these scenarios, the periocular region %\footnote{In the literature, the terms `ocular' and `periocular' are often used interchangeably. In this paper, we adopt the same convention.} 
offers a practical alternative for unconstrained biometrics.
In loose terms, it refers to the externally visible region of the face that surrounds the eye socket, encompassing the iris, sclera, eyelids, eyelashes, eyebrows, and surrounding skin (Fig.~\ref{fig:eye_anatomy}).
While iris recognition typically requires controlled near-infrared (NIR) acquisition at close range~\cite{[Nigam15]}, and face recognition performance degrades under unconstrained conditions \cite{2026PAMI_50yrfacerecognition_jain}, the periocular region can be acquired across a wider range of distances and imaging conditions. Moreover, it often remains visible when the face is partially occluded (e.g., in selfie or in-the-wild imagery) and can still provide useful biometric information when the iris cannot be captured at sufficient resolution due to standoff distance.
It appears in both iris and face sensor outputs, and does not require precise segmentation, unlike iris systems. 
%which remains visible in many occlusion scenarios and can be captured across a broad range of distances, sensors, and acquisition conditions.
%
%
This flexibility, convenience, and robustness of the periocular region to a wide range of unconstrained situations make it relevant for a variety of recognition tasks. 
Crucially for forensics or disinformation-detection applications, it retains soft-biometric cues (gender, age, ethnicity), as confirmed by recent surveys on periocular biometrics \cite{Alonso24computers_periSOA,zanlorensi22_AIR_ocular_db_competitions_survey,Kumari22_jksu_periocular_survey,sharma23cviu_periocular_masks_survey}. 
It has also been shown that the periocular region carries more soft-biometrics information than the iris texture alone \cite{KuehlkampBowyer19wacvGenderIrisHarder}.
%
%However, none of these surveys focuses specifically on demographic estimation from periocular imagery.
%

Soft-biometric attributes refer to ancillary, descriptive characteristics such as age, gender, ethnicity, height, hair colour, and other physical traits \cite{[Dantcheva16softbio]}.
Among them, demographic indicators (gender, age, ethnicity) have been particularly studied due to their relative permanence and discriminative capability. 
In unconstrained scenarios, exact identity inference can be unreliable, whereas soft-biometric attributes may be estimated with greater robustness even from low-quality and partial facial evidence typical of real-world surveillance and forensic scenarios \cite{Becerra2019airFaceSoftbioForensicSurveillSurvey}.  
Although demographic indicators do not uniquely identify an individual, they provide a complementary layer of \textit{soft} identity, with direct applicability in forensic and disinformation-detection applications, including:
(i) narrowing investigative search spaces, for example by filtering individuals in surveillance footage based on specific attributes \cite{Dantcheva11a}; 
(ii) supporting age-related forensic analysis and access-control scenarios, particularly in investigations involving minors; or 
(iii) identifying demographic inconsistencies in synthetic or manipulated media, given that generative models are known to struggle with demographic coherence across frames and facial regions \cite{2026PAMI_50yrfacerecognition_jain}.
The latter is increasingly relevant in the context of AI-generated and manipulated media, where demographic attributes may be synthesised, altered, or rendered inconsistently across frames, facial regions, or modalities.

This survey focuses on gender, age, and ethnicity estimation from the periocular region, including images that may contain the iris but do not rely solely on the segmented iris ring.
Works based only on isolated iris texture are considered only when they also report periocular results.
Despite a growing body of work on periocular recognition \cite{Alonso24computers_periSOA,sharma23cviu_periocular_masks_survey,zanlorensi22_AIR_ocular_db_competitions_survey,Kumari22_jksu_periocular_survey,[Rattani17soaOcularVIS],[Alonso16],[Nigam15],[Santos13]}, to the best of our knowledge, this is the first survey dedicated to demographic estimation from periocular imagery and its relevance to multimedia forensics and disinformation detection.

\begin{figure}[t]
  \centering
  % ---------------------------------------------------------------
  % FIGURE SUGGESTION (replaces/adapts Fig. 2 of the journal paper):
  % Show only a reduced grid of representative database samples --
  % suggest keeping 2 examples per category (face / iris / periocular),
  % 6 images total arranged as a 2x3 grid, with category labels.
  % This is a clear visual reduction from the full Fig. 2, avoids
  % exact duplication, and stays within the page budget.
  % Label each image with database name and resolution.
  % Placeholder below.
  % ---------------------------------------------------------------
  \includegraphics[width=0.9\textwidth]{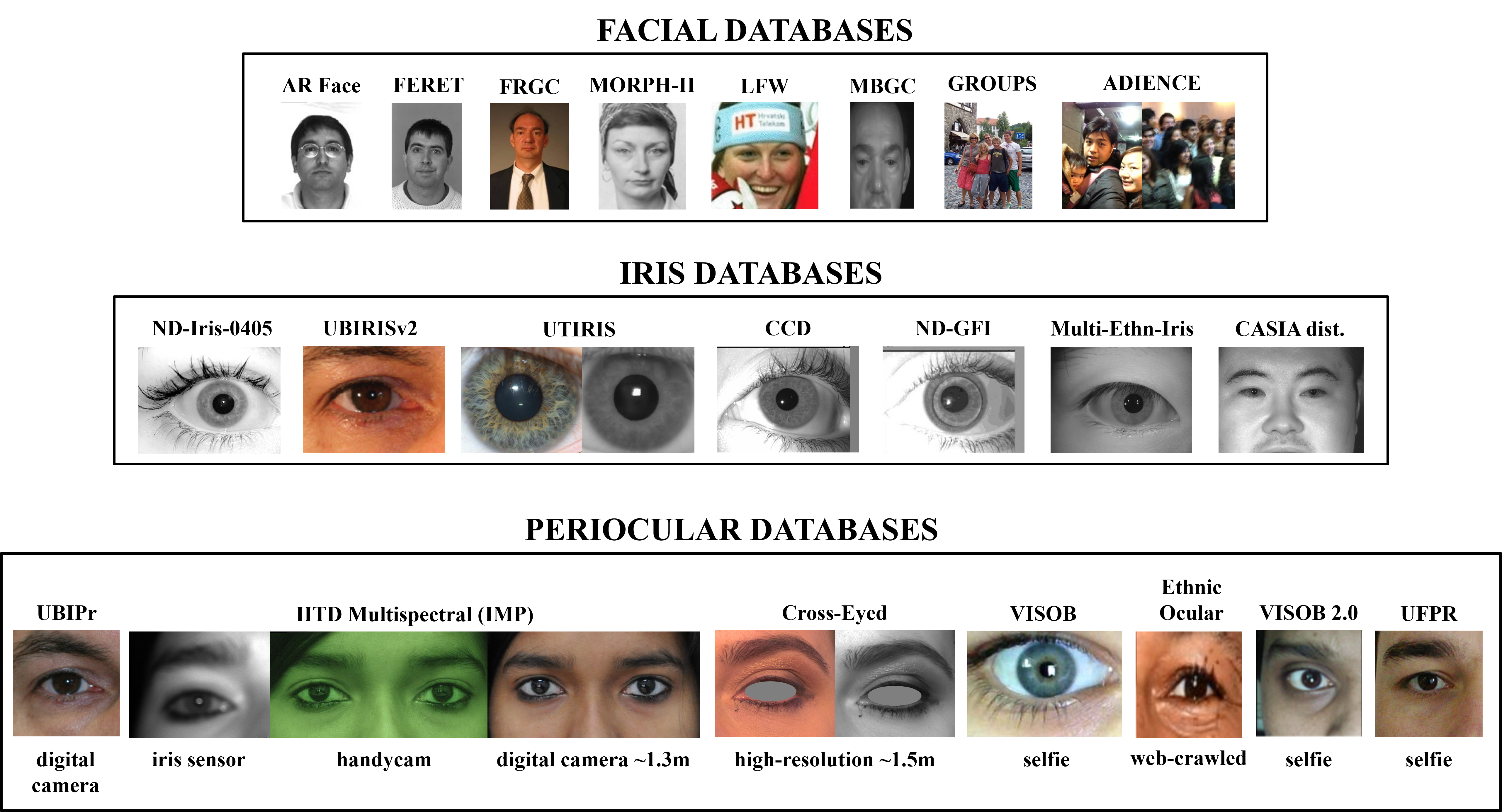}
  \caption{Representative samples from the three dataset categories (face/iris/periocular) used in periocular demographic research.}
  \label{fig:db_samples}
\end{figure}

\begin{table*}[!t]

\centering

%{|p{2.8cm}|p{3.6cm}|p{2.5cm}|p{1cm}|p{1.9cm}|p{1cm}|p{0.7cm}|p{1cm}|}

\resizebox{\textwidth}{!}{

%\begin{tabular}{|p{3.7cm}|p{2.2cm}|p{0.9cm}|p{1.1cm}|p{1.9cm}|p{0.7cm}|p{0.7cm}|p{0.1cm}p{0.1cm}p{0.1cm}|p{0.7cm}|p{0.7cm}|}

%\begin{tabular}{|c|c|c|c|c|p{0.7cm}|p{0.7cm}|p{0.1cm}p{0.1cm}p{0.1cm}|c|c|}

\begin{tabular}{cccc|ccc|ccc}

\toprule

\textbf{} & \textbf{} & \textbf{} & \multicolumn{1}{c|}{} & \textbf{} & \textbf{} & \multicolumn{1}{c|}{} & \multicolumn{3}{c}{\textbf{Best accuracy}}  \\ 

\textbf{Ref.} & \textbf{Year} & \textbf{Database} & \multicolumn{1}{c|}{\textbf{Type}} & \multicolumn{1}{c}{\textbf{Images}}  & \textbf{Subj.} & \multicolumn{1}{c|}{\textbf{Spectrum}}  & \multicolumn{1}{c}{\textbf{Gender}} &  \textbf{Age} &  \textbf{Ethnicity}  \\ 

\midrule

\cite{[Martinez98]}  &  1998  &  AR & face & +4000 & 126 & VIS & 85.61\% & - & -  \\ 

\cite{[Phillips00]}  &  2000  &  FERET & face & 14126 & 1199 & VIS & 90\% & - & 98.5\%  \\ 

\cite{[Phillips05]}  &  2005  &  FRGC & face & 36818 & 741 & VIS & 97.5\% & - & 98.7\%  \\ 
    
\cite{[Ricanek06]}  &  2006  &  MORPH-II  & face & 55k & 13k & VIS & - & 3.63 yr & - \\ 

\cite{[Huang07]}     &  2007  &  LFW  & face & 13233 & 5749 & VIS & 93.4\% & 60.2\% & 82.9\% \\ 
  
\cite{[Phillips09]}  &  2009  &  MBGC portal & face & 628 (V) & 114 & NIR & 98.2\% & - & 90\% \\ 

\cite{Gallagher09GROUPSfacedb}  &  2009  &  GROUPS  & face & 28231 & n/a & VIS & 87.37\% & - & - \\ 

\cite{[EidingerHassner14_Adience]}  &  2014  &  Adience  & face & 26580 & 2284 & VIS & 84.06\% & 51.03\% & - \\ 

\midrule

\cite{[Proenca10ubirisv2]}  &  2010  &  UBIRISv2 & iris & 11102 & 261 & VIS & 97.6\% & -  & -  \\ 

\cite{Doyle13BTAScontactlens_detect_NDdb}  &  2013  &  CCD & iris & 5100 & n/a  & NIR & 77.8\% & - & - \\ 

%\cite{[Phillips10]}
\cite{bowyer2016ndiris0405irisimagedataset} &  2016  &  ND-Iris-0405 & iris & 64980 & 356 & NIR & 82.53\% & - & 94.33\% \\ 

\cite{Hosseini10TIM_UTIRISdb}  & 2016   &  UTIRIS  & iris & 1540 & 79 & V+N & 86.89\% & - & -  \\ 

\cite{Bobeldyk16biosigIrisOcularGenderNIR}  &  2016  &  BioCOP  & iris & 43810 & 1096 & NIR & 85.9\% & - & 90.1\%  \\ 

\cite{Tapia16tifsGenderIrisCode}  &  2016  &  ND-GFI  & iris & 3000 & 1500 & NIR & 89.22\% & - & -  \\ 

\cite{[Singh17ijcbGenderRaceNIRIris]}  &  2017  &  Multi-Ethn-Iris & iris & 60310 & n/a & NIR & - & - & 97.38\% \\ 
 
\cite{KuehlkampBowyer19wacvGenderIrisHarder}  & 2019   &  ND-GFI-C  & iris & 6240 & 2005 & NIR & 80.8\% & - & - \\ \
  
\cite{[Viedma19ietGenderPeriocularNIR]} &  2019  &  UNAB  & iris & 2768 & 135 & NIR & 82.33\% & - & - \\ 

\cite{[CASIAirisdb]}  &  -  &  CASIA-Iris-Dist & iris & 2567 & 142 & NIR & 78\% & - & -  \\  

\midrule

\cite{[Padole12]}   &  2012  &  UBIPr & periocular & 10950 & 261  & VIS & 99.67\% & - & - \\ 

%\cite{[Sequeira14MobBIO]}  &  MobBIO  & periocular & 1680 & 105 & VIS & $\square$ & - & - & - & - \\ \hline

\cite{[Sharma14]}  &  2014  &  IMP  & periocular & 1240 & 62 & multi & 99.14\% & -  & - \\ 

%\cite{[Santos14]}  &  CSIP  & periocular & 2004 & 50 & VIS & $\square$ & - & - & - & - \\ \hline
  
%\cite{[Marsico15]}  &  MICHE-I  & periocular & 3732 & 92 & VIS & $\square$ & - & - & - & - \\ \hline

\cite{[sequeira16crosseyed]}  & 2016   &  Cross-Eyed & periocular & 3840 & 120 & V+N & 90\% & - & - \\ 
  
\cite{Rattani2016icipVISOB}  &  2016  &  VISOB  & periocular & 158136 & 550 & VIS & 90.2\% & - & -  \\ 

\cite{2019_ICB_periocular_LBCP_dual_CNN_Ethnic_Ocular_db_Tiong}  & 2019   &  Ethnic-Ocular  & periocular & 85394 & 1034 & VIS & 94.14\% & - & - \\ 

\cite{Nguyen21_icip_visob2}  &  2021  &  VISOB 2.0  & periocular & n/a & 250 & VIS & 84.89\% & - & -  \\ 

\cite{zanlorensi2022_SR_UFPR_db}  &  2022  &  UFPR   & periocular & 33660 & 1122 & VIS & 99.68\% & - & - \\ 

\bottomrule

\\

\end{tabular}

}

\caption{\label{tab:databases}
Publicly available datasets used in periocular demographic research.  
%G=Gender, A=Age, E=Ethnicity.  
VIS=Visible spectrum. NIR=Near-infrared spectrum.
V+N=VIS+NIR.
(V)=Videos. 
Best accuracy columns give the top reported value in the literature (Tables~\ref{tab:gender_soa} and~\ref{tab:age_ethnicity_soa}).
}

\end{table*}

%% ===========================================================
\section{Datasets for Periocular Demographic Estimation}
\label{sec:datasets}
%% ===========================================================
 
Table~\ref{tab:databases} summarises publicly available datasets used in periocular demographic research, while representative samples are shown in Figure~\ref{fig:db_samples}.  
The datasets can be grouped into three main categories:

\begin{itemize}
    \item Face databases, commonly used by cropping the periocular region from full-face images. They are usually captured in the visible (VIS) spectrum with conventional cameras and often provide relatively high-resolution imagery. Some of them (LFW, GROUPS, and Adience) are collected from online sources such as Flickr, and given their uncontrolled nature, these images exhibit greater variability in scale, pose, illumination, and image quality. The MBGC database is the only NIR dataset in this group, consisting of subjects walking naturally through a portal.

    \item Iris databases, typically acquired in the near-infrared (NIR) spectrum using dedicated close-up sensors under controlled acquisition conditions. Such images often include part of the surrounding periocular area, although with limited spatial extent. %Although originally designed for iris recognition, they have also been reused in periocular research, although the available area around the eye can be limited in some cases.
    The exception to close-up acquisition is CASIA-Iris-Distance, which was captured at a distance of 3 meters using a long-range NIR camera. In addition, UTIRIS contains images in both the NIR and VIS domains.

    \item Dedicated periocular databases. Most of them are captured in the VIS spectrum using smartphones or digital cameras, often under unconstrained but semi-controlled conditions, such as selfie acquisition or subjects looking at the device while an operator captures the image. Cross-Eyed contains both NIR and VIS images, whereas IMP is a multispectral database with images acquired under NIR, VIS and night-vision conditions.
    
\end{itemize}

As it can be seen, only a small proportion of databases are specifically designed for periocular research, and the majority are captured in controlled or semi-controlled conditions, with stationary subjects. 
In addition, very few databases contain multi-spectral data.

\subsection{Forensic Relevance of Datasets}

The datasets most representative of uncontrolled acquisition conditions are LFW, GROUPS, and Adience, which consist of images crawled from online sources and, therefore, reflect a wider range of in-the-wild variability. 
Although they mostly correspond to portrait-like or posed subjects, this type of imagery is still relevant for forensic investigations, since digital evidence often includes images collected from mobile devices, social media platforms, or online sources.
However, these datasets do not explicitly model other important forensic scenarios involving moving or non-cooperative subjects.
The MBGC database, where subjects are captured while walking through a biometric portal, is the only database containing moving subjects. Nevertheless, it is still a cooperative and dedicated acquisition setup based on NIR imaging, and therefore only partially reflects operational forensic conditions, where evidence is typically acquired with uncontrolled visible-spectrum cameras.
Thus, while portrait and selfie-like unconstrained imagery has received some attention, public periocular soft-biometrics databases still do not adequately cover more challenging forensic evidence such as CCTV footage, large stand-off distances, low resolution, non-cooperative subjects, motion blur, or compression artefacts.
Consequently, current benchmarks may not accurately reflect performance under such unconstrained forensic scenarios, potentially overestimating the accuracy expected in real forensic applications.

Periocular databases captured with digital cameras or smartphones in selfie or portrait mode are also representative of the type of digital evidence commonly found on mobile devices and social media platforms.
These datasets reflect visible-spectrum consumer imagery, including variations in pose, illumination, focus, camera quality, and user interaction.
Nevertheless, they are intentionally captured by cooperative users rather than obtained from incidental, degraded, or adversarial recordings.
Compared to LFW, GROUPS, or Adience, existing dedicated periocular databases are typically captured at fixed, controlled stand-off distances.
They also miss processing through social media pipelines, such as beautification filters, or compression \cite{Hedman22_pr_selfie_beauty_filters}. 
In a different direction, MORPH-II provides controlled mugshot-style imagery, which is also forensically relevant because it resembles law-enforcement records rather than incidental multimedia evidence.
Lastly, it must be highlighted that existing soft-biometrics research has not accounted for emerging multimedia forensic challenges such as AI-generated faces, deepfakes, or demographic attribute manipulation.

\subsection{Performance Trends Across Datasets}

Table~\ref{tab:databases} also provides the best accuracy reported in the literature for gender, age, and ethnicity estimation tasks. 
These results should be taken as an approximate indication, since different works may employ different subsets of the database or a different protocol.
An initial observation is that gender is, by far, the most extensively studied demographic attribute. 
Gender estimation generally achieves the highest performance, particularly on face and periocular datasets acquired in the visible spectrum, where accuracies exceed 95\% in several cases.
In contrast, NIR iris datasets tend to yield lower results despite their controlled acquisition at a very short stand-off distance. This suggests that skin texture and pigmentation cues available in VIS imagery play an important role. The much better results with UBIRISv2, an iris database but in the VIS spectrum, seem to confirm this observation. 
Ethnicity estimation also reports high accuracies, although conclusions are limited by severe class imbalance and inconsistent ethnicity definitions across datasets, as we will discuss later.
Age estimation remains the most challenging and least explored task, with substantially lower performance on unconstrained datasets such as LFW or Adience, highlighting the difficulty of extracting age-related cues from periocular imagery.

\begin{table}[t]

\centering

\resizebox{0.9\textwidth}{!}{%

%\begin{tabular}{|p{2.8cm}|p{2cm}|p{1.4cm}|p{0.7cm}|p{0.7cm}|p{0.6cm}|p{1cm}|p{1.3cm}|p{0.7cm}|p{2.2cm}|}
\begin{tabular}{p{0.7cm}|p{0.8cm}|p{5.2cm}|p{2.5cm}|p{1cm}|p{1cm}|p{1.1cm}}

\multicolumn{7}{c}{\textbf{PERIOCULAR GENDER ESTIMATION}} \\

\toprule

\textbf{Ref.} & \textbf{Year}  & \textbf{Method} & \textbf{Database} & \textbf{Spec-trum}  & \textbf{Eyes} & \textbf{Accu-racy} \\ \hline

\cite{[Kumari12]}  & 2012 &      ICA + NN & FERET & VIS & Both &  90\% \\ \hline

\cite{[Rattani17_GenderOcularHST]}  & 2017 &    (LBP, LTP, LPQ, BSIF, HOG) \newline + (SVM, MLP) & VISOB & VIS & One & 90.2\% \\ \hline

\cite{[Singh17ijcbGenderRaceNIRIris]}  & 2017 &    Deep Class-Encoder   & ND-Iris-0405  & NIR &  One &  82.53\% \\ \hline

\cite{[Tapia17ijcbGenderMultispectralOcular]}  & 2017 &   (Intensity, Texture, Shape) + RF & Cross-Eyed & V+N & One & 90\% \\ \hline

\cite{[Bhattacharyya19scGenderFacialRegionsGA]}  & 2019 &    compass LBP + SVM & Adience %, cFERET, LFW, CUFS, CUFSF  
& VIS & One \newline  Both & 84.06\% \newline  83.27\% \\ \hline

\cite{[BobeldykRoss19accessGenderRAceNIRocular]}  & 2019 &    (BSIF, LBP, LPQ) + \newline SVM & BioCOP\newline CCD-II  & NIR \newline NIR & One \newline One & 85.9\% \newline 77.8\%  \\ \hline 

\cite{Eskandari19ietGenderFaceOcular} & 2019 &    (ULBP+BSA) + SVM & CASIA-Iris-Dist \newline   & NIR\newline  &  One \newline Both & 66\% \newline 78\%  \\ \hline

\cite{2019_JISE_ocular_gender_multiple_models_Kao}  & 2019 &    PCA+RF & AR\newline  & VIS \newline & One \newline Both & 83.11\% \newline 85.61\% \\ \hline

\cite{KuehlkampBowyer19wacvGenderIrisHarder}  & 2019 &   (Location, Intensity) + SVM, \newline CNN+SVM, CNN &  ND-GFI-C & NIR & One & 80.8\% \\ \hline

\cite{2019_PAA_ocular_gender_nir_cnn_Manyala}  & 2019 &    CNN + SVM, CNN  & MBGC   & NIR & One  & 98.2\%  \\ \hline

\cite{[Viedma19ietGenderPeriocularNIR]}  & 2019 &    (Intensity, Texture, Shape) \newline + SVM, ensembles & ND-GFI\newline UTIRIS\newline UNAB & NIR \newline V+N \newline NIR & One  \newline One \newline One & 89.22\% \newline 86.89\% \newline 82.33\% \\ \hline

\cite{[Alonso20SoftBio]}  & 2020 &   CNN + SVM &  LFW & VIS & One \newline Both & 92.6\% \newline 93.4\% \\ \hline

\cite{KrishnanRattani2020icprFairnessMobileOcularGenderVISOB2}  & 2020 &   CNN &  VISOB 2.0 & VIS & One & 84.89\% \\ \hline

\cite{Talreja22_wacv_periocular_biometrics_leveraging_softbio}  & 2022 &    CNN & FRGC \newline UBIRISv2 & VIS \newline VIS & One \newline One &  97.5\% \newline 97.6\% \\ \hline

\cite{2023_IASC_ocular_gender_multiscale_Alrabiah}  & 2023 &    CNN & GROUPS \newline Ethnic-Ocular \newline IMP & VIS \newline VIS \newline NIR  &  One  \newline One \newline Both  &  87.37\% \newline 94.14\% \newline 99.14\%  \\ \hline

\cite{2025_MTA_cross_eyed_db_attention_ocular_gender_Kumar}  & 2025 &    CNN & UBIPr & VIS  & One  &  99.67\% \\ \hline

\cite{2025_SciReports_DL_ocular_ID_gender_Suravapu}  & 2025 &   ViT & UFPR & VIS  & One  &  99.68\% \\ 

\bottomrule 

\multicolumn{7}{c}{} \\

\end{tabular}
}

\caption{\label{tab:gender_soa}Selected results for periocular gender estimation (best results of each database). VIS=Visible spectrum. NIR=Near-infrared spectrum. V+N=VIS+NIR.}

\end{table}

%% ===========================================================
\section{Demographic Estimation: Methods and Results}
\label{sec:methods}
%% ===========================================================
 
We organise the methodological review by demographic attribute, providing a condensed overview of the main methodological trends and representative results. 
The best results reported in the literature on each database are summarized in Tables~\ref{tab:gender_soa} and~\ref{tab:age_ethnicity_soa}.
The review presented is derived from a comprehensive analysis of the periocular soft-biometrics literature. 
For conciseness and due to space limitations, we report only the best-performing result found for each database and demographic attribute. 
While this inevitably leaves out many relevant studies, the broader survey was used to identify overall trends, methodological developments, and open research challenges.

\subsection{Gender Estimation}
\label{sec:gender}

Gender is the most studied demographic attribute in periocular soft biometrics, typically formulated as a binary classification.
The literature shows a clear methodological progression from handcrafted descriptors and classical classifiers to convolutional neural networks, transfer learning, and, more recently, transformer-based models (Table~\ref{tab:gender_soa}), although transformer-based approaches remain comparatively scarce.

\paragraph{\textbf{Handcrafted descriptors.}}
\textbf{  }
%\noindent\textbf{Handcrafted features.}

\noindent Foundational works can be traced back to 2010, e.g. \cite{[Lyle10],[Merkow10]}, where it was common to gather data by cropping the periocular region from well-established face recognition databases.
Over the years, typical texture descriptors such as Local Binary Patterns (LBP), Histograms of Oriented Gradients (HOG), Binarized Statistical Image Feature (BSIF), Local Phase Quantization (LPQ), etc., have been employed, coupled with classifiers such as Support Vector Machines (SVM) or Random Forests (RF) \cite{[Rattani17_GenderOcularHST],[Tapia17ijcbGenderMultispectralOcular],[Bhattacharyya19scGenderFacialRegionsGA],[BobeldykRoss19accessGenderRAceNIRocular],Eskandari19ietGenderFaceOcular,KuehlkampBowyer19wacvGenderIrisHarder,[Viedma19ietGenderPeriocularNIR]}.

These works already demonstrated that periocular cues can support gender prediction, especially on visible-spectrum data, with accuracies exceeding 90\% on periocular databases such as VISOB \cite{[Rattani17_GenderOcularHST]} or the multi-spectral Cross-Eyed database \cite{[Tapia17ijcbGenderMultispectralOcular]}.
%%%%%%%%%%%%%%%
%results handcrafted
%%%%%%%%%%%%%%%
%face
%adience 84.06
%%%%%%%%%%%%%%%
%iris
%biocop 85.9
%ccd-ii 77.8
%casia dist 78
%nd-gfi-c 80.8
%nd-gfi 89.22
%utiris 86.89
%unab 82.33
%%%%%%%%%%%%%%%
%periocular
%VISOB 90.2
%cross eyed 90
%%%%%%%%%%%%%%%
%
%
On the other hand, experiments on the web-crawled Adience dataset have yielded lower accuracy (84.06\%) \cite{[Bhattacharyya19scGenderFacialRegionsGA]}, highlighting the difficulty of demographic estimation under unconstrained conditions.
Eyebrow shape features \cite{[Dong11ijcbEyebrowGender]} also proved effective, achieving 97\% accuracy on the FRGC database, close to the best result of 97.5\% reported in Table~\ref{tab:gender_soa}. This demonstrates the discriminative power of this sub-region, a finding repeatedly confirmed by subsequent works \cite{[Tapia17ijcbGenderMultispectralOcular],[Viedma19ietGenderPeriocularNIR]}.

A key finding from this period concerns the relative roles of the iris vs. the periocular area. 
Bobeldyk and Ross \cite{Bobeldyk16biosigIrisOcularGenderNIR} showed that restricting the analysis to the iris region alone reduces gender classification accuracy below 75\%, compared with 85\% when using the full periocular region on the BioCOP NIR database.
Kuehlkamp and Bowyer \cite{KuehlkampBowyer19wacvGenderIrisHarder} later quantified that periocular predictions are, on average, 17\% more accurate than predictions based on normalized iris texture.
These results suggest that gender-related cues are not confined to the iris texture itself but extend to the surrounding periocular region, including eyelids, eyebrows, and periocular skin.
The work \cite{KuehlkampBowyer19wacvGenderIrisHarder} also identified cosmetics as a confounding factor. The authors noted that it is difficult to determine whether a classifier learns gender-related cues from the intrinsic iris/periocular texture or instead exploits the presence of makeup. This effect was supported by the observed decrease in accuracy when images with cosmetics were removed from the training set.

Bobeldyk and Ross \cite{BobeldykRoss2019wacvwSoftbioOcular30pixels} further studied the effect of image down-sampling. Using BSIF+SVM and a small two-layer CNN on the BioCOP database, they reported a gender accuracy of 85.4\%. More importantly, they showed that demographic cues remain detectable even at extremely low resolutions, achieving 77.7\% accuracy with images of only $10\times12$ pixels and 77.6\% with images of $5\times6$ pixels.

A few works have exploited the availability of databases captured in several spectra simultaneously, such as Cross-Eyed \cite{[Tapia17ijcbGenderMultispectralOcular]} or UTIRIS \cite{[Viedma19ietGenderPeriocularNIR]}. 
These studies showed that combining VIS and NIR information can improve performance. 
This is relevant for forensic investigations, where evidence may originate from heterogeneous sensors and acquisition conditions, including not only conventional visible-spectrum cameras but also night-vision CCTV or infrared-illuminated surveillance systems. 
However, while these works demonstrated the feasibility of periocular gender estimation across heterogeneous spectra, they also indicated that robust cross-domain generalisation remains challenging.

\paragraph{\textbf{Deep learning descriptors.}}
\textbf{  }

\noindent CNN-based approaches have progressively replaced handcrafted features. 
To avoid overfitting, early works employed small custom CNNs \cite{[Rattani18_GenderOcularIETB],[Viedma18ipasGenderOcularNIR],2023_IJMCE_gender_ocular_EfficientNet_Nambiar}, networks pretrained on ImageNet \cite{[Viedma18ipasGenderOcularNIR],[Alonso20SoftBio],2023_IJCCE_ocular_gender_transfer_learning_Kumar,2023_IJMCE_gender_ocular_EfficientNet_Nambiar}, mobile architectures \cite{[Alonso21ietSoftBio]} or face recognition backbones such as VGG-Face \cite{2019_JISE_ocular_gender_multiple_models_Kao}.
Semi-supervised learning with large amounts of unlabeled data has also been proposed to cope with the limited availability of large-scale annotated datasets \cite{2021_SSCI_Unlabeled_data_gender_age_face_ocular_Nadimpalli_Rattani}.

The use of deep-learned features has pushed gender classification accuracy above 99\% on dedicated periocular datasets such as UFPR and UBIPr. 
However, another selfie database, VISOB 2.0, still reports an accuracy of 84.89\%.
VISOB 2.0 is a database that, compared to UFPR and UBIPr, exhibits a greater variability in cameras and illumination, suggesting that differences in capture conditions, acquisition devices, illumination, evaluation protocol, or other covariates may affect performance and remain an open issue for further investigation.
Web-crawled databases such as LFW and GROUPS also report lower accuracies (93.4\% and 87.37\%, respectively), reflecting the increased difficulty of in-the-wild imagery.

A recent line of work has explored the joint training of identity recognition and soft-biometric classification through multi-task learning \cite{Talreja22_wacv_periocular_biometrics_leveraging_softbio}. Joint training of the two tasks promotes mutual regularisation and complementary representations for both identity and soft-biometrics, improving the overall discriminative power of the model. Using this approach, the authors reported the highest accuracies in the reviewed literature on the FRGC and UBIRISv2 databases.

\begin{table}[t]

\centering

\resizebox{0.96\textwidth}{!}{%

%\begin{tabular}{p{0.7cm}|p{5.2cm}|p{2.5cm}|p{1cm}|p{1cm}|p{1.1cm}}

\begin{tabular}{|p{0.7cm}|p{0.8cm}|p{4.2cm}|p{2.4cm}|p{1cm}|p{2.1cm}|p{1cm}|p{2.2cm}|}

\multicolumn{8}{c}{\textbf{PERIOCULAR AGE ESTIMATION}} \\

\toprule

\textbf{Ref.} & \textbf{Year} & \textbf{Method} & \textbf{Database} & \textbf{Spec-trum} & \textbf{Classes} &  \textbf{Eyes} & \textbf{Accuracy} \\ \hline

\cite{[Yi14accvAgeFaceCNN]}  &  2014  &  23 sub-CNNs &   MORPH-II & VIS & 62 & Face \newline parts & 3.63 yr (MAE) \\ \hline

\cite{[Angeloni19iccvwAgeFaceParts]}  & 2019  &  4 sub-CNNs &  Adience & VIS & 8 & Face \newline parts & 51.03\% %±4.63 
(group) \newline 
83.41\% %±3.17 
(1-off) \\ \hline

\cite{[Alonso20SoftBio]}  &  2020  &  CNN + SVM &  LFW & VIS & 3 & One \newline Both & 60.2\% (group) \newline 60.0\% (group) \\ 

\bottomrule 

\multicolumn{8}{c}{} \\

%\multicolumn{10}{c}{} \\

\multicolumn{8}{c}{\textbf{PERIOCULAR ETHNICITY ESTIMATION}} \\

\toprule

\textbf{Ref.} & \textbf{Year} & \textbf{Method} & \textbf{Database} & \textbf{Spec-trum} & \textbf{Classes} &  \textbf{Eyes} & \textbf{Accuracy} \\ \hline

\cite{[Lyle12]} &  2012 &  (LBP, HOG, DCT, LCH) \newline + (SVM, ANN) & MBGC & NIR & AS, NAS & One  &  90\% \\ \hline 

\cite{Mohammad2017ceecOcularEthnicity} & 2017  &   (LBP, HOG) \newline + (SVM, MLP, LDA, QDA) & FERET & VIS  & ME, NME & Both  &  98.5\% \\ \hline 

\cite{[Singh17ijcbGenderRaceNIRIris]}  &  2017  &   Deep Class-Encoder   &  ND-Iris-0405  & NIR &  CA, AS & One & 94.33\%  \\ \cline{4-7}
  &   &  &  Multi-Ethn-Iris  & NIR &  CA, CH, IN & One & 97.38\% \\ \hline

\cite{[BobeldykRoss19accessGenderRAceNIRocular]}  &  2019 &   (BSIF, LBP, LPQ) \newline + SVM & BioCOP  & NIR & CA, OT & One & 90.1\% \\ \hline 

\cite{[Alonso20SoftBio]}  &  2020  &   CNN + SVM &  LFW & VIS & WH, BL, AS, \newline IN, OT & One \newline Both & 82.9 \newline 81.3\% \\ \hline

\cite{Talreja22_wacv_periocular_biometrics_leveraging_softbio}  & 2022  &  CNN &  FRGC & VIS & n/a  & One  &  98.7\% \\ 

\bottomrule 

\multicolumn{8}{c}{} \\

\end{tabular}
}

\caption{\label{tab:age_ethnicity_soa}Selected results for periocular age and ethnicity estimation (best results of each database). VIS=Visible spectrum. NIR=Near-infrared spectrum. MAE=Mean Absolute Error in years. AS=Asian, NAS=no-Asian, ME=Middle-East, NME=no-Middle-East, CA=Caucasian, CH=Chinese, IN=Indian, WH=White, BL=Black, OT=Other. 
%HI=Hispanic, NA=Native-American, AF=African
The name of the ethnic classes are as defined by the authors of the respective papers mentioned in the table.}

\end{table}

\subsection{Age Estimation}
\label{sec:age}

Age is considered as the most challenging demographic attribute to infer, even for humans, \cite{Carcagni2015jivpDemographicsDifferentConfigs}, because it varies continuously and it is influenced by a plethora of internal and external factors related with biology, genetics, lifestyle, health, climate, environment, etc. In addition, men and women may have different appearance at the same age, as well as people from diverse ethnic groups \cite{Becerra2019airFaceSoftbioForensicSurveillSurvey,[Sun18pamiDemographicsBiometricsSurvey]}.
The task is either formulated as regression (predicting exact age, evaluated by Mean Absolute Error, MAE) or as group classification (exact group or 1-off accuracy, the latter considering succesful also the groups adjacent to the true age group).
Different to the other demographics, age estimation also has an intrinsic \textit{order relationship} between age values or groups.

Comparatively, periocular age estimation is the most under-researched task of the three (gender, age, ethnicity).
Early works already relied on CNN-based features, with virtually no prior use of handcrafted descriptors, highlighting the relative novelty of the task compared with other periocular soft-biometric attributes.
The most employed public database is Adience, which contains 8 age classes. The database is provided with a multi-fold evaluation protocol, making comparison between different works possible. 
The best published results in the literature \cite{[Angeloni19iccvwAgeFaceParts]} with Adience are 51.03\% %±4.63 
(exact) and 83.41\% %±3.17 
(1-off).
The work \cite{[Alonso20SoftBio]} made use of the LFW database, which contains annotations of 5 age groups, but given the under-representation of children of very young age, the authors used only 3 groups corresponding to ''minors'', ''adults'' and ''seniors'', achieving an accuracy of 60.2\% on the exact group estimation.

Both Adience and LFW are web-crawled datasets exhibiting substantial in-the-wild variability, which may partly explain the lower performance observed for periocular age estimation compared with gender estimation.
The MORPH-II database, on the other hand, contains controlled mugshot-style images with exact age labels ranging from 16 to 77 years old. The best results reported on this more controlled database, formulated as a regression task, is a Mean Absolute Error (MAE) of 3.63 years \cite{[Yi14accvAgeFaceCNN]}, showing that acquisition quality and controlled imaging conditions have a strong impact on periocular age estimation.

A recent work on a self-captured database (thus not shown in Table~\ref{tab:age_ethnicity_soa}) addressed pediatric age estimation from NIR images of children aged 4-16 years, achieving MAE=1.33 years with periocular inputs and a group accuracy of 83.82\% using two age groups (4-9 years old, early chilhood, and 10-16 years old, early adolescence) \cite{2025_Access_iris_ocular_pediatric_age_Venkataswamy_Schuckers}. 
This is the first dedicated work on this sub-problem, with direct relevance to child-safety forensic applications. 
The authors also observed a performance drop to MAE=2.03 years and 72.45\% group accuracy when only iris inputs were used. 
This behaviour is consistent with previous findings in gender estimation, where the full periocular region also outperformed the isolated iris texture \cite{Bobeldyk16biosigIrisOcularGenderNIR,KuehlkampBowyer19wacvGenderIrisHarder}.
Taken together, these results suggest that demographic cues are not confined to the iris, but are strongly present in the surrounding periocular region.

\subsection{Ethnicity Estimation}
\label{sec:ethnicity}

Ethnicity estimation from periocular images has attracted more attention than age estimation, but remains understudied relative to gender estimation.
Ethnicity estimation studies suffer from two important structural limitations: 
(1)~inconsistent class definitions across datasets, since no standard taxonomy of ethnicities exists \cite{Fu14pamiRaceFaceSurvey}; and
(2)~severe class imbalance, with some ethnic groups being strongly under-represented in existing databases.
This is because many of the databases were not originally collected for ethnicity estimation but for other purposes, such as face recognition, and were later annotated with ethnicity labels. 
As a result, most works use only two ethnic groups, often reflecting the dominant demographic composition of the region where the database was captured, rather than the diversity required for robust real-world deployment.

Existing ethnicity estimation works (Table~\ref{tab:age_ethnicity_soa}, bottom) have either employed crops from face databases such as MBGC, FERET, LFW or FRGC, or NIR iris databases such as ND-Iris-0405, Multi-Ethnicity-Iris, or BioCOP. 
The use of dedicated periocular databases remains largely absent from the ethnicity estimation literature. 
Early works using handcrafted features 
%
%on face database crops, restricting experiments to two classes~\cite{[Lyle10],[Lyle12],Mohammad2017ceecOcularEthnicity}.
%
reported high accuracies, 
namely 90\% on MBGC (NIR data) for Asian vs. no-Asian classification \cite{[Lyle12]} and
98.5\% on FERET (VIS data) for Middle-East vs. non-Middle-East classification \cite{Mohammad2017ceecOcularEthnicity}.
On NIR iris data of BioCOP, the work \cite{[BobeldykRoss19accessGenderRAceNIRocular]} reported an accuracy of 90.1\% using handcrafted features for Caucasian vs. Other classification.

The use of deep-learned features pushed the accuracy on the FRGC database to 98.7\% \cite{Talreja22_wacv_periocular_biometrics_leveraging_softbio}. 
In another work \cite{[Singh17ijcbGenderRaceNIRIris]}, the accuracy on NIR iris data for three-class classification with the Multi-Ethnicity-Iris database was 97.38\%, and 94.33\% on ND-Iris-0405.
On the other hand, the more challenging web-crawled LFW dataset achieves 82.9\% accuracy \cite{[Alonso20SoftBio]}, although the authors employed five classes, compared with two in the majority of the literature. Therefore, this lower accuracy should be interpreted carefully, as it may reflect both the greater difficulty of in-the-wild imagery and the greater complexity of a five-class ethnicity classification problem.

%\subsection{Commonalities and Limitations of Existing Studies}

% ============================================================
\section{Limitations and Open Challenges}
\label{sec:challenges}

Despite the advances reviewed in this paper, periocular soft-biometrics remains affected by several limitations and research challenges that we summarise here.

\paragraph{Class imbalance.} This is a recurrent limitation, since no database was specifically designed for balanced demographic analysis. 
The issue is especially problematic in age and ethnicity studies. %, where classes are consistently under-represented.  
In age estimation, most databases are strongly concentrated in adult groups, with limited representation of children or elderly subjects. For example, MORPH-II is dominated by subjects between 20 and 49 years old, while LFW contains mostly adult and senior faces.
Adience provides broader coverage of younger age groups, but its distribution remains uneven, with the largest group at ages 25--32 and relatively few samples above 48 years old.
A similar issue appears in ethnicity estimation, where several datasets are dominated by one or two groups, such as Caucasian subjects in FERET, FRGC, LFW, ND-Iris-0405, and BioCOP (above 62\% in all cases), or Indian subjects in Multi-Ethnicity-Iris (68.8\%). This imbalance complicates the interpretation of reported accuracies and comparisons across studies, since high overall performance may mask poor performance for underrepresented demographic groups.
Future datasets should therefore be designed with demographic balance as an explicit criterion. %, and evaluations should report per-group performance in addition to overall accuracy.

\paragraph{Standardisation and benchmarking.} 
Since most databases were not collected for demographic estimation, they often lack standardised soft-biometrics evaluation protocols. This makes direct comparison across papers difficult, since different works may use different subsets, train/test partitions, or even closed-set protocols in which the same users appear in both the training and test sets.
A more systematic effort is thus needed to develop standard benchmark databases and open-set evaluation protocols for periocular soft biometrics, enabling comparison and ensuring proper generalisation to unseen populations.
The problem of standardisation is also evident in ethnicity estimation, where the absence of a standard taxonomy of ethnic categories further limits comparability across datasets and studies. 
Thus, future work would benefit from a definition of standard ethnicity categories. %, together with benchmark datasets that provide balanced class distributions and fixed evaluation protocols.

\paragraph{Fairness, bias, and ethical concerns.}
Fairness concerns in periocular soft-biometrics are closely linked to the class imbalance and lack of standard benchmarks discussed above, in line with broader concerns about demographic bias in biometric systems \cite{Drozdowski20TTS_DemographicBiasSurvey}.
If datasets are demographically imbalanced and evaluation protocols are not standardised, models trained on these datasets may propagate or amplify existing biases when deployed in operational settings.
Under these conditions, it also becomes difficult to determine whether reported performance is consistent across demographic groups or whether high overall accuracy hides systematic errors for underrepresented populations.
This raises ethical concerns in contexts such as forensic investigation, where demographic prediction errors may have disproportionate consequences for specific groups.
Thus, in addition to balanced datasets and standardised class definitions and evaluation protocols, future work should move beyond aggregated accuracy and include fairness-aware benchmarks with per-group performance reporting.

\paragraph{Controlled vs.\ in-the-wild gap.} 
Most existing databases are collected in controlled or semi-controlled environments, often with stationary, cooperative subjects.
Under these conditions, the literature reports very high performance, with accuracies approaching 99\% in some gender or ethnicity estimation works, and Mean Absolute Errors as low as 3.63 years for age estimation.
However, results on web-crawled in-the-wild databases such as LFW, GROUPS, and Adience are consistently lower than those obtained on more controlled datasets across the three demographic tasks (Tables~\ref{tab:gender_soa} and~\ref{tab:age_ethnicity_soa}).
Although these datasets reflect forms of digital evidence commonly found on mobile devices, social media platforms, or online sources, they still do not cover more challenging forensic conditions, such as degraded CCTV footage, moving subjects, distant captures without user cooperation, compression artefacts, or processing through social media pipelines.
Moreover, no dedicated periocular soft-biometrics benchmark currently includes synthetic or manipulated imagery, limiting the evaluation of periocular demographic inference under synthetic media or manipulated content.

\paragraph{Cross-database, cross-sensor and cross-spectral generalisation.}
A few databases, such as UTIRIS, BioCOP, IMP, Cross-Eyed, VISOB, VISOB 2.0, and UFPR, have been collected with multiple sensors or spectral bands. 
These resources enable evaluation of the robustness of periocular soft-biometric methods across acquisition devices and spectral domains.
On the other hand, most works employ a single database in their experiments, with only a few assessing generalisation by testing a classifier trained on a different database \cite{[BobeldykRoss19accessGenderRAceNIRocular]} or by training on multiple datasets in different spectra \cite{Tapia18sitisSexOcularMultispectra}.
As a result, research on domain shift in periocular soft biometrics remains limited, particularly regarding how reliably demographic cues transfer across sensors, spectra, and databases.
This is especially relevant for practical deployment, where models should generalise across domains, a robustness not yet systematically studied.

% ============================================================
\section{Forensic and Disinformation Detection Applications}
\label{sec:forensics}

We discuss in this section potential applications of periocular demographic estimation in forensic and disinformation detection contexts.

\paragraph{Subject search-space reduction}
A canonical use case for soft biometrics is candidate filtering. 
Rather than identifying an individual, soft-biometric attributes can be used to reduce the search space by filtering out candidates whose demographic or visual attributes are inconsistent with those of the person of interest, especially if full identity recognition is unreliable \cite{[Dantcheva16softbio]}.
This is particularly relevant in large-scale image, video or biometric databases, where demographic attributes can support subject retrieval, suspect filtering, or candidate-list prioritisation \cite{[Sun18pamiDemographicsBiometricsSurvey]}.
The periocular region is particularly valuable in unconstrained contexts \cite{Becerra2019airFaceSoftbioForensicSurveillSurvey}, where it may be the only consistently visible area of the face, either involuntarily, due to occlusions by environmental elements, clothing, masks, work gear, or scarves, or voluntarily, when offenders attempt to conceal their faces \cite{sharma23cviu_periocular_masks_survey}.

\paragraph{Child protection and age verification}
Age estimation is relevant to child-safety applications. 
Existing age-estimation methods often focus on exact age prediction or multi-class age-group classification.
However, in scenarios involving children, the goal is often not to determine the exact chronological age of a subject, but to support threshold-based triage decisions, such as whether the subject is likely to fall below a legally or operationally defined age threshold.
From this perspective, coarse threshold cases may be more operationally relevant than fine-grained age estimation.
%
%This use case is consistent with broader applications of soft biometrics in age-specific access control, security monitoring, and multimedia retrieval \cite{[Dantcheva16softbio],[Sun18pamiDemographicsBiometricsSurvey]}.
%
In forensic investigations, such estimates could help prioritise large volumes of visual evidence by flagging material that requires expert review \cite{[Macedo18sibgraphiChildPornDetectionBenchmark]}.
Here, periocular analysis is particularly relevant, since the eye region may be the only visible facial area in partially occluded, cropped, or degraded imagery. 
Recent pediatric work has shown that age estimation from NIR periocular images is feasible in children aged 4--16 years using a database of 21,000 images from 288 children \cite{2025_Access_iris_ocular_pediatric_age_Venkataswamy_Schuckers}.  
However, these results should be interpreted as evidence of feasibility rather than operational readiness. 
The authors acknowledge the controlled nature of the acquisition setup using close-up iris sensors and highlight the need to assess more realistic conditions, including variable lighting, motion artefacts, reduced subject cooperation, and lower image quality.  

\paragraph{Synthetic data and soft biometrics consistency.}
The use of periocular soft biometric consistency cues for supporting the analysis of AI-generated or manipulated media remains an emerging, largely unexplored research direction.
Generative methods can synthesise photorealistic faces in which the apparent demographics of the periocular region may be inconsistent with other facial features \cite{2026PAMI_50yrfacerecognition_jain}.
However, multi-region demographic consistency checking has received little attention, despite its relevance to disinformation detection and synthetic-media analysis
Detecting inconsistencies in demographic attributes estimated from the periocular region vs. the full face could provide a complementary cue for identifying synthetic or manipulated images.
In continuous video, temporal instability in soft biometric predictions across frames can also provide evidence of synthetic or altered faces, for example, when estimated age, gender, or ethnicity cues fluctuate in ways that are unlikely under natural video acquisition.

%% ===========================================================
\section{Conclusion}
\label{sec:conclusion}
%% ===========================================================

This paper surveyed demographic attribute estimation from periocular images, focusing on gender, age, and ethnicity prediction. 
We reviewed public datasets, methodological trends from handcrafted descriptors to deep learning and transformer based models, and the best reported results in the literature. 
Overall, the evidence confirms that the periocular region is a valuable source of soft-biometric information when full-face recognition is unreliable, particularly under partial occlusion, low image quality, or unconstrained acquisition conditions \cite{Alonso24computers_periSOA}.
%
%The periocular region is a valuable soft-biometric source under partial face occlusion \cite{Alonso24computers_periSOA}, with relevance to forensic and disinformation detection tasks such as subject search-space reduction, child protection, age verification, or consistency analysis to identify synthetic or manipulated multimedia content.
%

Gender estimation is the most extensively studied task and achieves the strongest results, with accuracies approaching 99.7\% on controlled or semi-controlled periocular datasets. 
However, performance drops on web-crawled and in-the-wild data, suggesting that controlled benchmarks may overestimate operational performance. 
Age remains the least explored and most challenging attribute, while ethnicity estimation is limited by inconsistent class definitions and severe demographic imbalance. 
Across all three tasks, current datasets still provide limited coverage of realistic forensic conditions, including degraded CCTV footage, moving subjects, distant capture, non-cooperation, compression artefacts, social-media processing, and synthetic or manipulated imagery. 
Other unresolved limitations include class imbalance, lack of common benchmarks, limited open-set evaluation, and insufficient cross-database, cross-sensor, and cross-spectral testing to ensure proper generalization.

Future work should prioritise demographically balanced benchmarks, standardised protocols, subject-disjoint and open-set partitions, per-group performance reporting, and systematic cross-domain evaluation. 
For forensic and disinformation detection applications, datasets should also better represent realistic evidence and include synthetic or manipulated faces. 
Methodologically, transformer-based architectures and large pretrained vision or vision-language foundation models offer promising directions for transfer learning, zero-shot or few-shot, and robust feature extraction \cite{2025_OtroshiTIFS_Foundation_Models_Biometrics_Survey}. 
%
%However, their use should be accompanied by careful validation of domain shift, demographic bias, and failure cases before deployment in high-stakes forensic or surveillance scenarios.
%
However, their use in periocular softbiometrics research is scarce \cite{2023_ApplSci_Ocular_ID_gender_ViT_Suravarapu,2025_SciReports_DL_ocular_ID_gender_Suravapu}.

\section*{Acknowledgements}
This work was supported by the Swedish Research Council (VR project 2021-05110) and by the EU Horizon Europe project PopEye (Grant Agreement No.~101168317). Funded by the EU. Views and opinions are those of the authors only and do not reflect those of the EU or the European Research Executive Agency.

\bibliographystyle{splncs04}
%\bibliography{main}

%\bibliography{fernando1}

\end{document}